\documentclass[journal]{IEEEtran}
\usepackage{cite}

\usepackage{graphics}

\ifCLASSINFOpdf
   \usepackage[pdftex]{graphicx}
\else
  \usepackage[dvips]{graphicx}
    \usepackage{graphicx}
\fi
\usepackage[cmex10]{amsmath}
\usepackage{algorithmic}

\usepackage[pagebackref]{hyperref}

\ifCLASSOPTIONcompsoc
  \usepackage[caption=false,font=normalsize,labelfont=sf,textfont=sf]{subfig}
\else
  \usepackage[caption=false,font=footnotesize]{subfig}
\usepackage{url}

\begin{document}
%
\title{A Bayesian Framework for Sparse Representation-Based 3D Human Pose Estimation}
%
%
%

\author{Behnam~Babagholami-Mohamadabadi, 
        Amin~Jourabloo, 
        Ali~Zarghami, 
        and~Shohreh~Kasaei~\IEEEmembership{Senior Member,~IEEE}
\thanks{The authors are with the Department of Computer Engineering, Sharif University of Technology, Tehran, Iran, e-mail: skasaei@sharif.edu}

}

\maketitle

\begin{abstract}
A Bayesian framework for 3D human pose estimation from monocular images based on {\it sparse representation} (SR) is introduced. Our probabilistic approach aims at simultaneously learning two overcomplete dictionaries (one for the visual input space and the other for the pose space) with a shared sparse representation. Existing SR-based pose estimation approaches only offer a point estimation of the dictionary and the sparse codes. Therefore, they might be unreliable when the number of training examples is small. Our Bayesian framework estimates a posterior distribution for the sparse codes and the dictionaries from labeled training data. Hence, it is robust to overfitting on small-size training data.
Experimental results on various human activities show that the proposed method is superior to the state-of-the-art pose estimation algorithms.
\end{abstract}


%
\IEEEpeerreviewmaketitle
\section{Introduction}
%
%
 \IEEEPARstart{R}{ecently}, 3D human pose estimation from monocular images has attracted much attention in computer vision community due to its significant role in various applications; such as visual surveillance, activity recognition, motion capturing, etc. Although many algorithms have been proposed  for estimating the 3D human poses from single images, it has been remained as a challenging task due to the lack of depth information and significant variations in visual appearances, hunam shapes, lightning conditions, clutters, and the forth.\\
Existing methods for monocular 3D pose estimation can be divided into three main categories. The \textit{model-based approaches} which assume a known parametric body model and estimate the human pose by inverting the kinematics or by optimizing an objective function of pose variables \cite{Agarwal,Andriluka}. These computationally expensive approaches need an accurate body model and a good initialization stage. Furthermore, due to non-convexity of their objective functions, their solution might be sub-optimal.
 On the other hand, the \textit{learning-based approaches} employ a direct mapping between the visual input space and the human pose space \cite{Elgammal,Lee}. Despite the superiority of these approaches, one of their major drawbacks is that their pose estimation accuracy depends on the amount of training data.
Finally, the  \textit{examplar-based approaches} estimate the pose of an unknown visual input (image) by searching the training data (a set of visual inputs whose corresponding 3D pose descriptors are known) and interpolating from the poses of similar training visual input(s) to the unknown visual input \cite{Jiang,Odobez}. The problem with these methods is that their computational complexity is high (because of searching the whole visual input database to find the similar sample(s) to an unknown input).\\
Some researchers have recently utilized the {\it sparce representation and dictionary learning} (SRDL) framework to estimate the human pose \cite{Huang, Huang2,Su}.
{\it Huang et~al.} \cite{Huang} proposed a SR-based method in which each test data point is expressed as a compact linear combination of training visual inputs. It is capable of dealing with occlusion. The pose of the test sample can be recovered by the same linear combination of the training poses. {\it Ji et~al.} \cite{Su} introduced a robust {\it dual dictionaries learning} (DDL) approach which can handle corrupted input images. An efficient algorithm is also provided to solve the DDL optimization model.\\
Although the results of SRDL approaches are comparable with the state-of-the-art methods, they suffer from two shortcomings. Firstly, since these algorithms only provide a point estimation of the dictionary and the sparse codes (which might be sensitive to the choice of training examples), they tend to overfit the training data (especially when the number of training examples is small). Secondly, none of these methods can use the information of the pose training data. Precisely speaking, all of the SRDL-based methods learn the dictionary and sparse codes without considering the fact that the dictionary should be learned such that the samples of similar poses have similar sparse codes. In order to overcome these shortcomings, this paper presents a Bayesian framework for SRDL-based pose estimation that targets the popular cases for which the number of training examples is limited. Moreover, by employing appropriate prior distributions on the latent variables of the proposed model, the dictionary is learnt with the constraint that the samples with similar poses must have similar sparse codes.\\
The remainder of this letter is organized as follows: The proposed method is introduced in Section \ref{pm}. Experimental results are presented in Section \ref{er}. Finally, the conclusion and future work are given in Section \ref{co}.
 \section{Proposed 3D Human Pose Estimation Method}\label{pm}
 Following \cite{Su}, we aim at learning two dictionaries (visual input dictionary and pose dictionary) with a shared sparse representation based on a Bayesian learning framework that utilizes the information of the pose training data. \\
 Let $X = \{\boldsymbol{x}_{i}\in \mathcal{R}^{M_{x}}\}_{i=1}^{N}$, and $Y =  \{\boldsymbol{y}_{i}\in \mathcal{R}^{M_{y}}\}_{i=1}^{N}$ denote the training set of $N$ visual input features and their corresponding pose features, respectively. We model each input feature $\boldsymbol{x}_{i} (i=1,...,N)$ and pose feature $\boldsymbol{y}_{i} (i=1,...,N)$ as a sparse combination of the atoms of dictionaries $D^{x}\in \mathcal{R}^{M_{x}\times K}$ and $D^{y}\in \mathcal{R}^{M_{y}\times K}$ with an additive noise $\boldsymbol{e}_{i}^{x}$ and $\boldsymbol{e}_{i}^{y}$, respectively.The matrix form of the model is given as
 \begin{equation}\label{model}
 X = D^{x}A + E^{x}, \;\; Y = D^{y}A + E^{y}
 \end{equation}
 where $A = [\boldsymbol{\alpha}_{1},\boldsymbol{\alpha}_{2},...,\boldsymbol{\alpha}_{N}] \in \mathcal{R}^{K\times N}$ is the set of $K$-dimensional sparse codes, $E^{x}\sim \mathcal{N}(0, \gamma_{xy}^{-1}I_{M_{x}})$, and $E^{y}\sim \mathcal{N}(0, \gamma_{xy}^{-1}I_{M_{y}})$ are the zero-mean Gaussian noise with precision value $\gamma_{xy}$ ($I_{M_{x}}$ and $I_{M_{y}}$ are $M_{x}\times M_{x}$ and $M_{y}\times M_{y}$ identity matrices, respectively). We model each sparse code $\boldsymbol{\alpha}_{i} = [\alpha_{1i},...,\alpha_{Ki}]^{T} (i=1,...,N)$ as an element-wise multiplication of a binary vector $\boldsymbol{z}_{i} = [z_{1i},...,z_{Ki}]^{T} (i=1,...,N)$ and a weight vector $\boldsymbol{s}_{i} = [s_{1i},...,s_{Ki}]^{T} (i=1,...,N)$, as
 \begin{equation}
\alpha_{ki} = \frac{z_{ki} + 1}{2} s_{ki},\;\; i=1,...,N,\;k=1,...,K
 \end{equation}
 where $\alpha_{ki}$ denotes the $k$-th coefficient of the $i$-th sparse code, $z_{ki}\in \{-1,1\} (i=1,...,N, k=1,...,K)$ is a binary random variable, and $s_{ki}  (i=1,...,N, k=1,...,K)$ is a zero-mean Gaussian random variable with precision value $\gamma_{s}$ $\big(s_{ki}\sim \mathcal{N}(0,\gamma_{s}^{-1})\big)$. The intuition for the above model is that if $z_{ki}=-1$, then $\alpha_{ki} = 0$ and the $k$-th atom of the dictionaries $D^{x}$ and $D^{y}$ are inactive. If $z_{ki}=+1$, then the $k$-th atom of the dictionaries are active, and the value of cofficient $\alpha_{ki}$ is drawn from $\alpha_{ki}\sim\mathcal{N}(0,\gamma_{s}^{-1})$. We also put a prior distribution on each binary random variable $z_{ki} (i=1,...,N, k=1,...,K)$ by using the logistic sigmoid function, as
 \begin{equation}\label{binary}
 P(z_{ki}\mid w_{ik}) = \frac{1}{1+e^{-z_{ki}w_{ik}}},\;i=1,...,N, k=1,...,K
 \end{equation}    
 where $\{\boldsymbol{w}_{k} = [w_{1k},...,w_{Nk}]^{T}\}_{k=1}^{K}$ are the hyper-parameters of the model. 
In order to exploit the information of the training pose data, a prior Gaussian distribution is considered for $\{\boldsymbol{w}_{k}\}_{k=1}^{K}$ as
\begin{equation}\label{sigma}
 \boldsymbol{w}_{k} \sim \mathcal{N}(0,\Sigma_{w}),\;\; k=1,...,K
 \end{equation} 
 where 
 \begin{equation}
 \Sigma_{w}(i,j) = \mathcal{K}(\boldsymbol{y}_{i},\boldsymbol{y}_{j}),\;i=1,...,N, j=1,...,N
 \end{equation}
 and $\mathcal{K}(\boldsymbol{y}_{i},\boldsymbol{y}_{j})$ is a valid kernel (a kernel which satisfies the Mercer's condition) that diminishes by increasing the distance between $\boldsymbol{y}_{i}$ and $\boldsymbol{y}_{j}$. 
By using above distributions, the process of generating the sparse codes is as follows:
 \begin{itemize}
 \item Draw the parameters $\{\boldsymbol{w}_{k}\}_{k=1}^{K}$ by using (\ref{sigma}).
 \item Draw the binary random vectors $\{\boldsymbol{z}_{i}\}_{i=1}^{N}$ by using (\ref{binary}).
 \item If $z_{ki} = 1$, draw $\alpha_{ki}\sim \mathcal{N}(0,\gamma_{s}^{-1})$, else $\alpha_{ki} = 0$. 
 \end{itemize}
As it can be seen from this process, if the two input features have similar pose features, they tend to use the same dictionary atoms (imposed by kernel $\mathcal{K}$) to get similar sparse codes.\\
In our method, we also impose a prior zero-mean Gaussian distribution on the dictionary atoms of $D^{x} = [\boldsymbol{d}_{1}^{x},...,\boldsymbol{d}_{K}^{x}]$ and $D^{y}= [\boldsymbol{d}_{1}^{y},...,\boldsymbol{d}_{K}^{y}]$, as
\begin{equation}
\boldsymbol{d}_{k}^{x} \sim \mathcal{N}(0,\gamma_{x}^{-1}I_{M_{x}}),\;\boldsymbol{d}_{k}^{y} \sim \mathcal{N}(0,\gamma_{y}^{-1}I_{M_{y}}),\;k=1,...,K.
\end{equation} 
To be Bayesian, we typically place non-informative Gamma hyper-priors on parameters $\gamma_{x},\gamma_{y},\gamma_{xy}$, and $\gamma_{s}$.
Given the training data $(X,Y)$, the proposed hierarchical probabilistic model can be expressed as
\begin{figure}[!t]
\centering
\includegraphics[width=3in,height=3.3in]{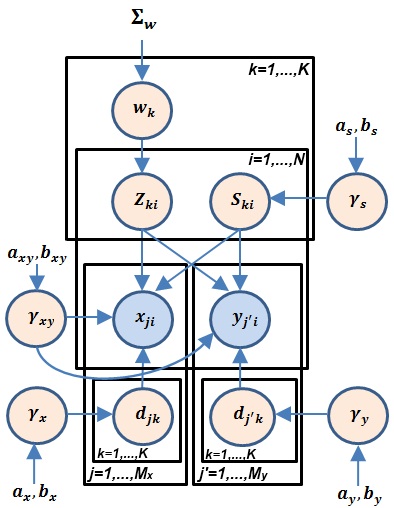}
\caption{Proposed model (blue shadings indicate observations).}
\label{fig_sim}
\end{figure}
\begin{equation}
P(W \mid \Sigma_{w}) \sim \prod_{k=1}^{K}\mathcal{N}(\boldsymbol{w}_{k}; 0, \Sigma_{w})
\end{equation}
\begin{equation}\label{logistic}
P(Z\mid W) = \prod_{k=1}^{K}\prod_{i=1}^{N}\frac{1}{1+exp(-z_{ki}w_{ik})}
\end{equation}
\begin{equation}
P(S\mid \gamma_{s})\sim \prod_{k=1}^{K}\prod_{i=1}^{N}\mathcal{N}(s_{ki};0,\gamma_{s}^{-1})
\end{equation}
\begin{equation}
P(\gamma_{s}\mid a_{s},b_{s})\sim Gamma(\gamma_{s};a_{s},b_{s})
\end{equation}
\begin{equation}
P(X\mid Z,S,D^{x},\gamma_{xy}) \sim \prod_{i=1}^{N}\mathcal{N}(\boldsymbol{x}_{i};D^{x}(\frac{\boldsymbol{z}_{i}+1}{2}\odot\boldsymbol{s}_{i}),\gamma_{xy}^{-1}I)
\end{equation}
\begin{equation}
P(Y\mid Z,S,D^{y},\gamma_{xy}) \sim \prod_{i=1}^{N}\mathcal{N}(y_{i};D^{y}(\frac{\boldsymbol{z}_{i}+1}{2}\odot\boldsymbol{s}_{i}),\gamma_{xy}^{-1}I)
\end{equation}
\begin{equation}
P(\gamma_{xy}\mid a_{xy},b_{xy})\sim Gamma(\gamma_{xy};a_{xy},b_{xy})
\end{equation}
\begin{equation}
P(D^{x}\mid \gamma_{x})\sim \prod_{j=1}^{M_{x}}\prod_{k=1}^{K}\mathcal{N}(d_{jk}^{x};0,\gamma_{x}^{-1})
\end{equation}
\begin{equation}
P(\gamma_{x}\mid a_{x},b_{x})\sim Gamma(\gamma_{x};a_{x},b_{x})
\end{equation}
\begin{equation}
P(D^{y}\mid \gamma_{y})\sim \prod_{j'=1}^{M_{y}}\prod_{k=1}^{K}\mathcal{N}(d_{j'k}^{y};0,\gamma_{y}^{-1})
\end{equation}
\begin{equation}
P(\gamma_{y}\mid a_{y},b_{y})\sim Gamma(\gamma_{y};a_{y},b_{y})
\end{equation}
where $\odot$ is the element-wise multiplication operator, $Z = [\boldsymbol{z}_{1},...,\boldsymbol{z}_{N}]$, $S = [\boldsymbol{s}_{1},...,\boldsymbol{s}_{N}]$, $W = [\boldsymbol{w}_{1},...,\boldsymbol{w}_{K}]$ are the hidden variables, $\Gamma = \{\gamma_{x},\gamma_{y}, \gamma_{s}, \gamma_{xy}\}$ are the parameters (the precision values, inverse variances, of the Guassian noise distributions), and $\Phi = \{a_{xy},b_{xy},a_{x},b_{x},a_{y},b_{y},a_{s},b_{s}\}$ are the hyper-parameters of the proposed model. The graphical representation of the proposed probabilistic model is shown in Fig \ref{fig_sim}.
\subsection{Posterior Inference}\label{PI}
Due to intractability of computing the exact posterior distribution of the hidden variables, the inference is performed by using the Gibbs sampling to approximate the
posterior with $S$ samples. In the proposed model, all of the distributions are in the conjugate exponential form except for the logistic function. Due to the non-conjugacy between the logistic function and the Gaussian distribution, deriving the Gibbs update equation for $W$ in closed-form is intractable. To overcome this problem, one can put an exponential upper bound on the logistic functions of (\ref{logistic}) based on the convex duality theorem \cite{Jordan}. By using this theorem and utilizing the fact that the log of a logistic function is concave, an upper bound on the logistic functions is obtained in the form of
\begin{equation}\label{bound}
\frac{1}{1+exp(-z_{ki}w_{ik})} \leq exp\big(\lambda_{i} z_{ki}w_{ik} - g(\lambda_{i})\big)\;\;\; i=1,...,N
\end{equation}
where
\begin{equation}
g(\lambda_{i})  = -\lambda_{i} \log \lambda_{i}  - (1-\lambda_{i})\log (1 - \lambda_{i})\;\;\; i=1,...,N
\end{equation}
and $\{\lambda_{i}\}_{i=1}^{N}$ are the variational parameters which should be optimized to get the tightest bound. 
By using the upper bound of (\ref{bound}), we propose a Gaussian distribution as the  distribution $Q$ in a {\it Metropolis-Hastings} (MH) independence chain algorithm \cite{Hastings} and derive the posterior samples for $W$ by using this algorithm (the details of generating samples for $W$ and other hidden variables are available in Appendix A). 
\subsection{Pose Prediction}
After computing the posterior distribution of hidden variables, in order to determine the target pose $\boldsymbol{y}_{t}$ of a given test instance $\boldsymbol{x}_{t}$, given the test instance, the predictive distribution of the target pose is first computed  by integrating out the hidden variables as
\begin{align*}
&P(\boldsymbol{y}_{t} \mid \boldsymbol{x}_{t}, X,Y) = \nonumber \\
&\sum_{\boldsymbol{z}_{t}}\int P(\boldsymbol{y}_{t}, \boldsymbol{s}_{t}, \boldsymbol{z}_{t}, D^{x}, D^{y},\boldsymbol{w}_{t}, \gamma_{xy},\gamma_{s}\mid \boldsymbol{x}_{t}, X,Y)\times \nonumber \\
&\;\;\;\;\;\;\;\;\;\;\;\;\;\;\;\;\;\;\;\;\;\;\;\;\;\;\;\;\;\;\;\;\;\;\;\;d\boldsymbol{s}_{t} \;d\gamma_{xy}\;d\gamma_{s}\;dD^{x}\;dD^{y}\;d\boldsymbol{w}_{t}
\end{align*}
\begin{align}\label{aaaaa}
&\propto \sum_{\boldsymbol{z}_{t}}\int P(\boldsymbol{y}_{t}\mid \boldsymbol{s}_{t}, \boldsymbol{z}_{t}, D^{y},\gamma_{xy})P(\boldsymbol{x}_{t}\mid \boldsymbol{s}_{t}, \boldsymbol{z}_{t}, D^{x},\gamma_{xy})\times \nonumber \\
&T(\boldsymbol{z}_{t},\boldsymbol{s}_{t}, D^{x}, D^{y}, \boldsymbol{w}_{t}, \gamma_{xy},\gamma_{s})d\boldsymbol{s}_{t} \;d\gamma_{xy}\;d\gamma_{s}\;dD^{x}\;dD^{y}\;d\boldsymbol{w}_{t}
\end{align}
where 
\begin{align}
&T(\boldsymbol{z}_{t},\boldsymbol{s}_{t}, D^{x}, D^{y}, \boldsymbol{w}_{t}, \gamma_{xy},\gamma_{s}) = P(D^{x}\mid X,Y)P(D^{y}\mid X,Y) \nonumber \\
&P(\boldsymbol{z}_{t} \mid \boldsymbol{w}_{t})P(\boldsymbol{s}_{t}\mid \gamma_{s})P(\gamma_{s}\mid X,Y)P(\boldsymbol{w}_{t}\mid X,Y) P(\gamma_{xy}\mid X,Y)
\end{align}
and $\boldsymbol{w}_{t} = [w_{t1},...,w_{tK}]^{T}$.
The mean of this distribution is the target pose $\hat{\boldsymbol{y}}_{t}$ for $\boldsymbol{x}_{t}$. 
Since the expression of (\ref{aaaaa}) cannot be computed in a closed-form fashion, one can resort to the Monte Carlo sampling to approximate that expression. 
As such, the distribution $T(\boldsymbol{z}_{t},\boldsymbol{s}_{t}, D^{x}, D^{y}, \boldsymbol{w}_{t}, \gamma_{xy},\gamma_{s})$ with $L$ samples can be approximated as
\begin{equation}
P(\boldsymbol{y}_{t} \mid \boldsymbol{x}_{t}, X,Y) \approx \frac{1}{L} \sum_{l=1}^{L}\beta_{l}\mathcal{N}(\boldsymbol{y}_{t};(D^{y})^{l}(\frac{\boldsymbol{z}_{t}^{l}+1}{2}\odot \boldsymbol{s}_{t}^{l}), (\gamma_{xy}^{l})^{-1}I)
\end{equation}
\begin{align}
\beta_{l} &= P(\boldsymbol{x}_{t}\mid \boldsymbol{s}_{t}^{l}, \boldsymbol{z}_{t}^{l}, (D^{x})^{l},\gamma_{xy}^{l})\nonumber \\
&= \mathcal{N}(\boldsymbol{x}_{t};(D^{x})^{l}(\frac{\boldsymbol{z}_{t}^{l}+1}{2}\odot \boldsymbol{s}_{t}^{l}), (\gamma_{xy}^{l})^{-1}I),\;\; l=1,...,L,
\end{align}
where $r^{l}$ is the $l$-th sample of the hidden variable $r$. 
By using the fact that the sum of Gaussian distributions is still a Gaussian distribution,  $\hat{\boldsymbol{y}}_{t}$ is computed analytically by
\begin{equation}
\hat{\boldsymbol{y}}_{t} = \frac{\sum_{l=1}^{L}\beta_{l}(D^{y})^{l}(\frac{\boldsymbol{z}_{t}^{l}+1}{2}\odot \boldsymbol{s}_{t}^{l}) }{L\sum_{l=1}^{L}\gamma_{xy}^{l}}.
\end{equation}
Sampling from $T(\boldsymbol{z}_{t},\boldsymbol{s}_{t}, D^{x}, D^{y}, \boldsymbol{w}_{t}, \gamma_{xy},\gamma_{s})$ is straightforward (we use the posterior samples, see Section \ref{PI}). 
However, due to the fact that the true pose of the unknown visual input $\boldsymbol{x}_{t}$ in unknown, $P(\boldsymbol{w}_{t}\mid X,Y)$ cannot be obtained, and hence the posterior samples of $\boldsymbol{z}_{t}$ cannot be generated. 
To overcome this problem, the samples of $\boldsymbol{w}_{t}$ are derived based on the posterior samples $W$ as
\begin{equation}
w_{tk}^{l} = \frac{1}{j} \sum_{i=1}^{N}(w_{ik}^{l})^{\theta_{i}}\;\;k=1,...,K,\;l=1,...,L,
\end{equation}
where $\theta_{i}=1$ if $\boldsymbol{x}_{i}$ belongs to the $j$ nearest neighbors of $\boldsymbol{x}_{t}$, and otherwise $\theta_{k}=0$. 
Sample derivation of $\boldsymbol{w}_{t}$ is based on the fact that neighboring visual inputs are more likely to have similar poses. By using the above samples for $\boldsymbol{w}_{t}$, one can generate the samples from $P(\boldsymbol{z}_{t}\mid \boldsymbol{w}_{t})$ by using (\ref{binary}).
\begin{table*}[t]
\renewcommand{\arraystretch}{1.3}
\caption{Average error (in degrees) with standard deviation of different methods.}
\label{table_example}
\centering
\begin{tabular}{|c|c|c|c|c|c|c|}
\hline
Activity & Tr. $\#$ & RVM & TGP & SR & DDL & PM\\
\hline
	Acrobatics 	& 30 & 15.963 $\pm$ \small2.73 &  15.411 $\pm$ \small2.97 & 14.005 $\pm$ \small2.32 & 16.731 $\pm$ \small3.82 & \textbf{12.595} $\pm$ \small1.24\\
	Acrobatics 	 & 60 & 13.294 $\pm$ \small2.53 &  13.353 $\pm$ \small2.49 & 12.805 $\pm$ \small2.12 & 14.734 $\pm$ \small3.41 & \textbf{9.328} $\pm$ \small0.99\\
    Acrobatics   	 & 100 & 10.651 $\pm$ \small1.52 & 9.882 $\pm$ \small1.76 & 8.104 $\pm$ \small1.44 & 10.323 $\pm$ \small1.97 & \textbf{6.443} $\pm$ \small0.96\\
Acrobatics & 200 & 7.247 $\pm$ \small1.14 &   6.896 $\pm$ \small1.05 & 5.506 $\pm$ \small0.93 & 6.989 $\pm$ \small1.31 & \textbf{4.862} $\pm$ \small0.79\\
        
\hline
	Navigate	& 30 & 10.821 $\pm$ \small1.19 &  10.917 $\pm$ \small1.31 & 10.455 $\pm$ \small0.99 & 11.421 $\pm$ \small1.53 & \textbf{7.623} $\pm$ \small0.61\\
	Navigate	& 60 & 6.674 $\pm$ \small0.96 &  6.819 $\pm$ \small0.89 & 6.662 $\pm$ \small0.71 & 7.872 $\pm$ \small1.24 & \textbf{5.782} $\pm$ \small0.34\\
      Navigate  & 100 & 4.434 $\pm$ \small0.22 & 5.029 $\pm$ \small0.36 & 5.550 $\pm$ \small0.49 & 5.753 $\pm$ \small0.57 & \textbf{3.229} $\pm$ \small0.20\\
Navigate & 200 & 3.567 $\pm$ \small0.17 & 4.194 $\pm$ \small0.16 & 3.866 $\pm$ \small0.27 & 4.331 $\pm$ \small0.41 & \textbf{3.075} $\pm$ \small0.12\\
\hline
	Golf	& 30 & 14.514 $\pm$ \small2.88 &  14.728 $\pm$ \small2.02 & 13.909 $\pm$ \small1.82 & 15.688 $\pm$ \small2.93 & \textbf{8.241} $\pm$ \small1.13\\
	Golf	& 60 & 9.337 $\pm$ \small2.51 &  8.964 $\pm$ \small1.79 & 9.745 $\pm$ \small1.11 & 10.949 $\pm$ \small2.21 & \textbf{5.752} $\pm$ \small0.88\\
      Golf & 100 & 7.652 $\pm$ \small1.32 & 7.515 $\pm$ \small0.69 & 5.467 $\pm$ \small0.42 & 7.442 $\pm$ \small0.61 & \textbf{3.931} $\pm$ \small0.52\\
Golf & 200 & 5.220 $\pm$ \small1.48 & 5.333 $\pm$ \small0.74 & 4.535 $\pm$ \small0.50 & 5.273 $\pm$ \small0.57 & \textbf{3.034} $\pm$ \small0.37\\
\hline
\end{tabular}
\end{table*}

\section{Experimental Results}\label{er}
\begin{figure}
\centering
\includegraphics[width=2.9in,height=2.9in]{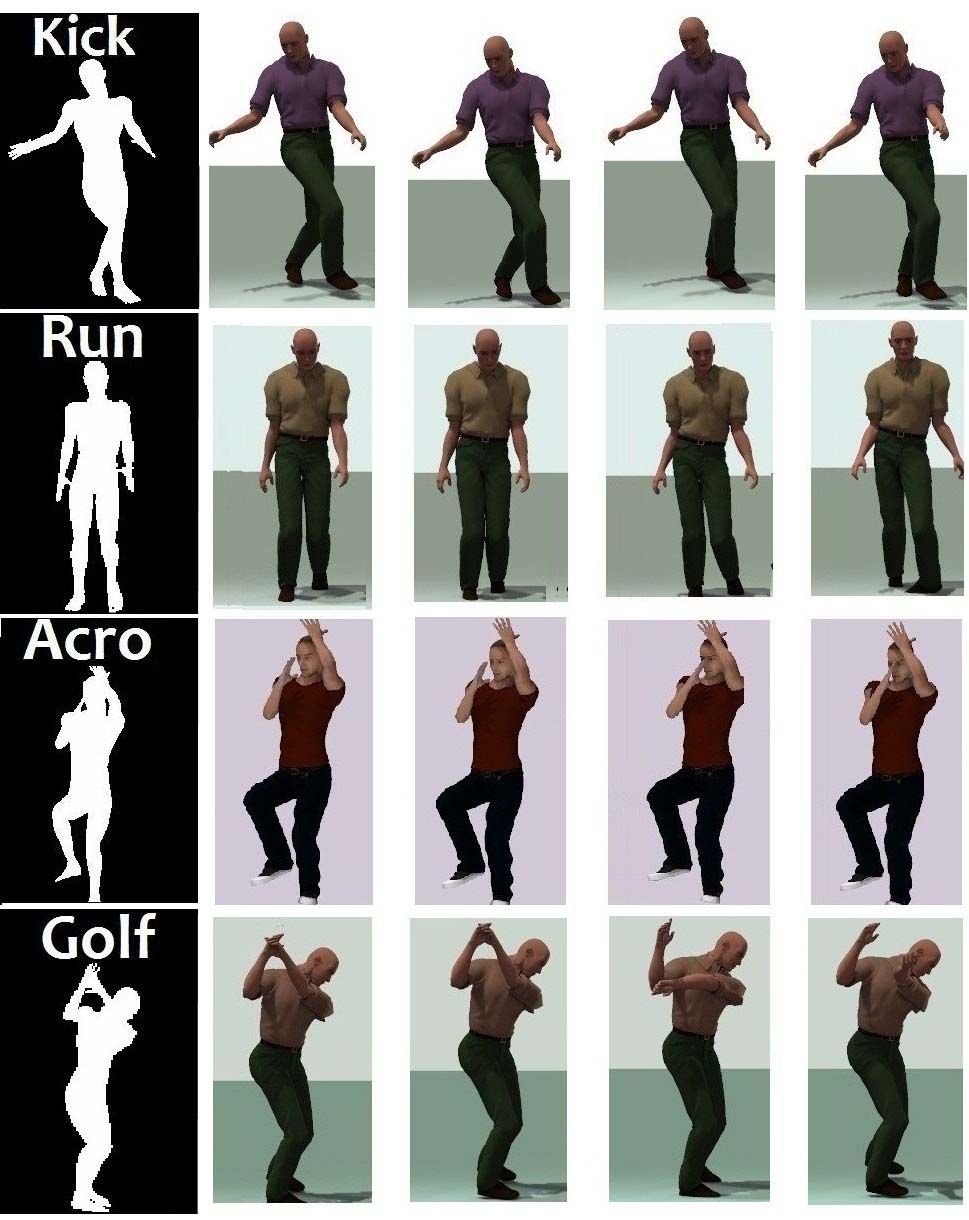}
\caption{Subjective comparison. Columns show input silhouettes, real labels, and outputs of PM, SR, and DDL, respectively.}
\label{subjective}
\end{figure}

In order to evaluate the performance of the proposed method, the activities in the CMU Mocap dataset\footnote{\href{http://mocap.cs.cmu.edu}{http://mocap.cs.cmu.edu/}} are used in the {\it bvh} format to generate the silhouettes of real sequences. The method is tested  on various activities ("Acrobatic", "Navigate", "Golf", etc). We have used the histograms of shape contexts \cite{Agarwal} which encodes the visual input (silhouette) into a 100-dimensional descriptor as the input feature. 
The human body pose is also encoded by 57 joint angles (three angles for each joint). The error is the average (over all angles) of the {\it root mean square error} (RMS). We captured 600 frames from each sequence and used 30, 60,100, and 200 of them as the training data, and the rest as the test data. In all experiments, 
all hyper-parameters are set to $10^{-6}$
to make the prior Gamma distributions uninformative. We also used the exponentional kernel $\mathcal{K}(\boldsymbol{y}_{i}, \boldsymbol{y}_{j}) = exp(-\|\boldsymbol{y}_{i}-\boldsymbol{y}_{j}\|/  \eta)$, for which the kernel parameter $\eta$ is set to
\begin{equation}
\eta = 2\sum_{i<j}\frac{\|\boldsymbol{y}_{i}-\boldsymbol{y}_{j}\|_{2}^{2}}{N(N-1)}.
\end{equation}
In order to determine an appropriate number of dictionary atoms. $K$, and nearest neighbors of unknown data samples, $j$, the five-fold cross validation approach is performed to find the best pair ($K$, $j$). The tested values for $K$ are $\{64, 128, 196, 256\}$ and for $j$ are $\{3, 5, 7\}$.
In the analysis that follows, 1200 MCMC iterations are used (700 burn-in and 500 collection, from a random start). 
For the proposal distribution $Q$ in MH algorithm, the acceptance rates were greater than $94\%$. 
We compared the performance of the proposed method with that of the {\it relevance vector machine} (RVM) as a well-known supervised regression method, the {\it twin Gaussian process} (TGP) \cite{Bo} as a state-of-the-art method, and DDL \cite{Su} and SR \cite{Huang2} as two state-of-the-art SRDL-based 3D human pose estimation methods. The average estimation accuracies (over 10 runs) together with the standard deviation for three activities are shown in Table \ref{table_example} (the results for other activities are available in Appendix B), from which we can see that the
proposed method significantly outperforms  the other methods. The improvement in performance is because of two reasons. Firstly,
the number of labeled data is small; hence these methods may overfit to the labeled data. Secondly, these methods cannot utilize the information of the pose data. Figure \ref{subjective} shows the subjective result of the proposed method, SR, and DDL for 4 sequences in the database, respectively. These outputs are obtained using 200 training data sampled from 400 test data. As can be seen, the proposed method has a better reconstruction rate than the other methods. 
 \section{Conclusion}\label{co}
In this letter, a fully probabilistic framework for SR-based 3D human pose estimation was proposed that utilized the information of the pose space. Experimental results proved the high performance of the proposed method especially in cases for which only a few number of training data is available.

\section*{Appendix A: MCMC Inference}
In the following equations, $P(q\mid -)$ denotes the
conditional probability of parameter $q$, given current value
of all other parameters. The sampling equations are as follows:
\subsection*{Sample $\boldsymbol{z}_{i}=[z_{1i},...,z_{ki},...,z_{Ki}]^{T}$:}
\begin{align}
P(z_{ki} \mid -) \propto &\prod_{j=1}^{M_{x}}\mathcal{N}(x_{ji};\sum_{k'=1}^{K}(d_{jk'}s_{k'i}\frac{z_{k'i}+1}{2}),\gamma_{xy}^{-1})\times\nonumber \\
&\prod_{j'=1}^{M_{y}}\mathcal{N}(y_{j'i};\sum_{k'=1}^{K}(d_{j'k'}s_{k'i}\frac{z_{k'i}+1}{2}),\gamma_{xy}^{-1})\times\nonumber \\
& \frac{1}{1+exp(-z_{ki}w_{ki})}
\end{align}
Since $z_{ki}$ is a Bernoulli random variable ($z_{ki}\in \{+1,-1\}$), we have
\begin{equation}
P(z_{ki}=+1 \mid -) \propto \alpha, \;\;\; P(z_{ki}=-1 \mid -) \propto \alpha'
\end{equation}
where
\begin{align}
\alpha = &exp\big(-\frac{1}{2}\gamma_{xy} \sum_{j=1}^{M_{x}}\big(x_{ji} - \sum_{k'=1}^{K}d_{jk'}s_{k'i}\big)^{2} \big)\times\nonumber \\
& exp\big(-\frac{1}{2}\gamma_{xy} \sum_{j'=1}^{M_{y}}\big(y_{j'i} - \sum_{k'=1}^{K}d_{j'k'}s_{k'i}\big)^{2} \big)\times\nonumber \\
& \frac{1}{1+exp(-w_{ki})}
\end{align}
\begin{align}
\alpha' = &exp\big(-\frac{1}{2}\gamma_{xy} \sum_{j=1}^{M_{x}}\big(x_{ji} - \sum_{k'\neq k}^{K}d_{jk'}s_{k'i}\big)^{2} \big)\times\nonumber \\
& exp\big(-\frac{1}{2}\gamma_{xy} \sum_{j'=1}^{M_{y}}\big(y_{j'i} - \sum_{k'\neq k}^{K}d_{j'k'}s_{k'i}\big)^{2} \big)\times\nonumber \\
& \frac{1}{1+exp(w_{ki})}
\end{align}
Hence, it is obvious that $z_{ki}$ is drawn from a Bernoulli distribution
\begin{equation}
P(z_{ki} \mid -) \sim Bernoulli(\theta), \;\;z_{ki}\in \{+1,-1\}
\end{equation}
where
\begin{equation}
\theta = \frac{\alpha}{\alpha + \alpha'}
\end{equation}
\subsection*{Sample $\boldsymbol{s}_{i}=[s_{1i},...,s_{ki},...,s_{Ki}]^{T}$:}
\begin{align}
P(s_{ki} \mid -) \propto &\prod_{j=1}^{M_{x}}\mathcal{N}(x_{ji};\sum_{k'=1}^{K}(d_{jk'}^{x}s_{k'i}\frac{z_{k'i}+1}{2}),\gamma_{xy}^{-1})\times\nonumber \\
&\prod_{j'=1}^{M_{y}}\mathcal{N}(y_{j'i};\sum_{k'=1}^{K}(d_{j'k'}^{y}s_{k'i}\frac{z_{k'i}+1}{2}),\gamma_{xy}^{-1})\times\nonumber \\
& \mathcal{N}(s_{ki};0,\gamma_{s}^{-1})
\end{align}
It is easy to show that if $z_{ki}=-1$, then $s_{ki}$ is drawn from 
\begin{equation}
P(s_{ki} \mid -)\sim \mathcal{N}(s_{ki};0,\gamma_{s}^{-1})
\end{equation}
and if $z_{ki}=1$, $s_{ki}$ is drawn from 
\begin{equation}
P(s_{ki} \mid -) \sim \mathcal{N}(\mu_{s},(\gamma _{s}')^{-1})
\end{equation}
where
\begin{equation}
\gamma _{s}' = \gamma_{s} + \big(\sum_{j=1}^{M_{x}}(d_{jk}^{x})^{2}+\sum_{j'=1}^{M_{y}}(d_{j'k}^{y})^{2}\big)\gamma_{xy}
\end{equation}

\begin{align}
\mu_{s} = (\gamma _{s}')^{-1}\gamma_{xy}&\bigg[\sum_{j=1}^{M_{x}}d_{jk}^{x}\big(\sum_{k'\neq k}^{K}(d_{jk'}^{x}s_{k'i}) - x_{ji}\big)\nonumber \\
& + \sum_{j'=1}^{M_{y}}d_{j'k}^{y}\big(\sum_{k'\neq k}^{K}(d_{j'k'}^{y}s_{k'i}) - y_{j'i}\big)\bigg]
\end{align}

\subsection*{Sample $W=[\boldsymbol{w}_{1},...,\boldsymbol{w}_{k},...,\boldsymbol{w}_{K}]$:}
\begin{align}\label{ww}
P(\boldsymbol{w}_{k}\mid -) &\propto \prod_{i=1}^{N} P(z_{ki}\mid w_{ik})\mathcal{N}(\boldsymbol{w}_{k};0,\Sigma_{w})\nonumber \\
 & \propto \prod_{i=1}^{N}\frac{1}{1+exp(-z_{ki}w_{ik})}\mathcal{N}(\boldsymbol{w}_{k};0,\Sigma_{w})
\end{align}
From the above equation, we can see that $P(\boldsymbol{w}_{k}\mid -)$ cannot directly sampled from. However, we can put an exponential upper bound on the logistic functions of the above equation based on the convex duality theorem \cite{Jordan}. Using this theorem and utilizing the fact that the log of a logistic function is concave, we obtain an upper bound on the logistic functions of the form
\begin{equation}\label{bound2}
\frac{1}{1+exp(-z_{ki}w_{ik})} \leq exp\big(\lambda_{i} z_{ki}w_{ik} - g(\lambda_{i})\big)\;\;\; i=1,...,N
\end{equation}
where
\begin{equation}
g(\lambda_{i})  = -\lambda_{i} \log \lambda_{i}  - (1-\lambda_{i})\log (1 - \lambda_{i})\;\;\; i=1,...,N
\end{equation}
$\lambda_{i} (i=1,...,N)$ are the variational parameters which should be optimized to get the tightest bound.\\
By substituting the above upper bound back into Eq. \ref{ww}, we obtain
\begin{equation}\label{normal}
P(\boldsymbol{w}_{k}\mid -) \leq \mathcal{N}(\boldsymbol{\mu}_{w},\Sigma_{w})
\end{equation} 
where
\begin{equation}
\boldsymbol{\mu}_{w} = \Sigma_{w}\Lambda_{k}
\end{equation}
\begin{equation}
\Lambda_{k} = [\lambda_{1}z_{k1},...,\lambda_{i} z_{ki},...,\lambda_{N}z_{kN}]^{T}
\end{equation}
We use this normal distribution (right-hand side of Eq. \ref{normal}) as the proposal distribution $Q$ in a 
Metropolis-Hastings (M-H) independence chain \cite{Hastings}, and accept $\boldsymbol{w}_{k}^{t+1} = \boldsymbol{w}_{k}'$ with probability $min\{p_{k},1\}$, where
\begin{align}
p_{k} &= \frac{P(\boldsymbol{w}_{k}')}{P(\boldsymbol{w}_{k}^{t})}\frac{Q(\boldsymbol{w}_{k}^{t})}{Q(\boldsymbol{w}_{k}')}\nonumber \\
&= \prod_{i=1}^{N}\frac{1+exp(-z_{ki}w_{ik}^{t})}{1+exp(-z_{ki}w_{ik}')}exp\big( -\boldsymbol{\mu}_{w}^{T}\Sigma_{w}^{-1}(\boldsymbol{w}_{k}' - \boldsymbol{w}_{k}^{t}) \big)
\end{align}
Since the proposal distribution $\big(\mathcal{N}(\boldsymbol{\mu}_{w},\Sigma_{w})\big)$ should be accurate around the current sample ($\boldsymbol{w}_{k}^{t}$), we can optimize the variational parameters to make the upper bound (right-hand side of Eq. \ref{bound2}) as tight as possible around the current sample. Hence, by replacing $w_{ik}$ with $w_{ik}^{t}$ in the right-hand side of Eq. \ref{bound2}, and by setting the derivative of the right-hand side of Eq. \ref{bound2} respect to $\{\lambda_{i}\}_{i=1}^{N}$ equal to zero, we can optimize $\{\lambda_{i}\}_{i=1}^{N}$ as
\begin{equation}
\lambda_{i} = \frac{1}{1+exp(-z_{ki}w_{ik}^{t})}\;\;\; i=1,...,N
\end{equation}
\subsection*{Sample $D^{x}=[\boldsymbol{d}_{1}^{x},...,\boldsymbol{d}_{k}^{x},...,\boldsymbol{d}_{K}^{x}]$:}
\begin{equation}
P(\boldsymbol{d}_{k}^{x}\mid -) \propto \prod_{i=1}^{N}\mathcal{N}(\boldsymbol{x}_{i};D^{x}(\frac{\boldsymbol{z}_{i}+1}{2}\odot \boldsymbol{s}_{i}),\gamma_{xy}^{-1}I)\mathcal{N}(\boldsymbol{d}_{k}^{x};0,\gamma_{x}I)
\end{equation}

It can be demonstrated that $\boldsymbol{d}_{k}^{x}$ is drawn from a Normal distribution

\begin{equation}
P(\boldsymbol{d}_{k}^{x}\mid -) \sim \mathcal{N}(\boldsymbol{\mu}_{k}^{x},\gamma_{x}'I)
\end{equation}
where
\begin{equation}
\gamma_{x}' = \big(\sum_{i=1}^{N}\frac{z_{ki}+1}{2}s_{ki}\big)^{2}\gamma_{xy} + \gamma_{x}
\end{equation}
\begin{align}
\boldsymbol{\mu}_{k}^{x} = &\frac{\sum_{i=1}^{N}(\frac{z_{ki}+1}{2}s_{ki})\gamma_{xy}}{\gamma_{x}'}I\times \nonumber \\
&\bigg( \sum_{i=1}^{N}\sum_{k'\neq k}^{K}\boldsymbol{d}_{k'}^{x}(\frac{z_{k'i}+1}{2}s_{k'i})
  - \boldsymbol{x}_{i}\bigg) 
\end{align}
\subsection*{Sample $D^{y}=[\boldsymbol{d}_{1}^{y},...,\boldsymbol{d}_{k}^{y},...,\boldsymbol{d}_{K}^{y}]$:}
\begin{equation}
P(\boldsymbol{d}_{k}^{y}\mid -) \propto \prod_{i=1}^{N}\mathcal{N}(\boldsymbol{y}_{i};D^{y}(\frac{\boldsymbol{z}_{i}+1}{2}\odot \boldsymbol{s}_{i}),\gamma_{xy}^{-1}I)\mathcal{N}(\boldsymbol{d}_{k}^{y};0,\gamma_{y}I)
\end{equation}

It can be demonstrated that $\boldsymbol{d}_{k}^{y}$ is drawn from a Normal distribution

\begin{equation}
P(\boldsymbol{d}_{k}^{y}\mid -) \sim \mathcal{N}(\boldsymbol{\mu}_{k}^{y},\gamma_{y}'I)
\end{equation}
where
\begin{equation}
\gamma_{y}' = \big(\sum_{i=1}^{N}\frac{z_{ki}+1}{2}s_{ki}\big)^{2}\gamma_{xy} + \gamma_{y}
\end{equation}
\begin{align}
\boldsymbol{\mu}_{k}^{y} = &\frac{\sum_{i=1}^{N}(\frac{z_{ki}+1}{2}s_{ki})\gamma_{xy}}{\gamma_{y}'}I\times \nonumber \\
&\bigg( \sum_{i=1}^{N}\sum_{k'\neq k}^{K}\boldsymbol{d}_{k'}^{y}(\frac{z_{k'i}+1}{2}s_{k'i})
  - \boldsymbol{y}_{i}\bigg) 
\end{align}
\subsection*{Sample $\gamma_{s}$:}
\begin{equation}
P(\gamma_{s}\mid -) \propto \prod_{k=1}^{K}\prod_{i=1}^{N}\mathcal{N}(s_{ki};0,\gamma_{s}^{-1})Gamma(\gamma_{s},a_{s},b_{s})
\end{equation}
It can be shown that $\gamma_{s}$ can be drawn from a Gamma distribution
\begin{equation}
P(\gamma_{s}\mid -) \sim Gamma(a'_{s},b'_{s})
\end{equation}
where,
\begin{equation}
a'_{s} = a_{s} + \frac{NK}{2},\;\;\; b'_{s} = b_{s} + \frac{1}{2}\sum_{k=1}^{K}\sum_{i=1}^{N}s_{ki}^{2}
\end{equation}
\subsection*{Sample $\gamma_{xy}$:}
\begin{align}
P(\gamma_{xy}\mid -) \propto &\prod_{j=1}^{M_{x}}\prod_{i=1}^{N}\mathcal{N}(x_{ji};\sum_{k=1}^{K}(d_{jk}^{x}\frac{z_{ki}+1}{2}s_{ki}),\gamma_{xy}^{-1})\times \nonumber \\
&\prod_{j'=1}^{M_{y}}\prod_{i=1}^{N}\mathcal{N}(y_{j'i};\sum_{k=1}^{K}(d_{j'k}^{y}\frac{z_{ki}+1}{2}s_{ki}),\gamma_{xy}^{-1})\times \nonumber \\
& \;\;\;\;\;\;\;\;\;\;\;\;Gamma(\gamma_{xy},a_{xy},b_{xy})
\end{align}
It is easy to show that $\gamma_{xy}$ can be drawn from a Gamma distribution
\begin{equation}
P(\gamma_{xy}\mid -) \sim Gamma(a'_{xy},b'_{xy})
\end{equation}
where,
\begin{equation}
a'_{xy} = a_{xy} + \frac{N(M_{x}+M_{y})}{2}
\end{equation}
\begin{align}
b'_{xy} = b_{xy} &+ \frac{1}{2}\sum_{j=1}^{M_{x}}\sum_{i=1}^{N}\big(x_{ji}-\sum_{k=1}^{K}(d_{jk}^{x}\frac{z_{ki}+1}{2}s_{ki})\big)^{2}\nonumber \\
& + \frac{1}{2}\sum_{j'=1}^{M_{y}}\sum_{i=1}^{N}\big(y_{j'i}-\sum_{k=1}^{K}(d_{j'k}^{y}\frac{z_{ki}+1}{2}s_{ki})\big)^{2}
\end{align}
\begin{table*}
\renewcommand{\arraystretch}{1.3}
\caption{Average error (in degrees) with standard deviation for different methods.}
\label{table_example2}
\centering
\begin{tabular}{|c|c|c|c|c|c|c|}
\hline
Activity & Tr. $\#$ & RVM & TGP & SR & DDL & PM\\
\hline
 Throw and catch football & 30 & 26.68 $\pm$ \small3.94 & 19.55 $\pm$ \small2.61 & 15.68 $\pm$ \small2.03 & 18.62 $\pm$ \small2.84 & \textbf{13.51} $\pm$ \small1.17\\
 Throw and catch football & 60 & 25.43 $\pm$ \small3.86 & 16.14 $\pm$ \small2.34 & 13.91 $\pm$ \small1.85 & 15.80 $\pm$ \small2.53 & \textbf{10.47} $\pm$ \small1.05\\
 Throw and catch football & 100 & 23.64 $\pm$ \small3.11  & 10.49 $\pm$ \small1.47 & 9.09 $\pm$ \small1.08 & 11.13 $\pm$ \small1.38 & \textbf{9.54} $\pm$ \small0.91\\
 Throw and catch football & 200 & 8.27 $\pm$ \small1.73 & 8.68 $\pm$ \small0.71 & 7.43 $\pm$ \small0.56 & 8.59 $\pm$ \small1.02 & \textbf{7.76} $\pm$ \small0.42\\
\hline
Michael jackson styled motions	 & 30 & 19.71 $\pm$ \small2.17 & 18.92 $\pm$ \small2.33 & 17.38 $\pm$ \small1.97 & 19.92 $\pm$ \small2.87 & \textbf{14.89} $\pm$ \small1.11\\
Michael jackson styled motions	 & 60 & 16.48 $\pm$ \small1.88 & 16.33 $\pm$ \small1.76 & 15.72 $\pm$ \small1.69 & 17.62 $\pm$ \small2.44 & \textbf{13.69} $\pm$ \small0.98\\
Michael jackson styled motions	 & 100 & 13.23 $\pm$ \small0.99 & 13.42 $\pm$ \small0.84 & 12.34 $\pm$ \small1.02 & 12.95 $\pm$ \small1.39 & \textbf{11.32} $\pm$ \small0.85\\
Michael jackson styled motions & 200 & 8.86 $\pm$ \small0.73 & 10.49 $\pm$ \small0.33 & 8.69 $\pm$ \small0.41 & 8.83 $\pm$ \small1.04 & \textbf{7.57} $\pm$ \small0.32 \\
\hline
kick soccer ball & 30 & 13.69 $\pm$ \small2.24 & 15.53 $\pm$ \small2.43 & 13.05 $\pm$ \small2.41 & 14.97 $\pm$ \small2.74 & \textbf{10.21} $\pm$ \small1.00\\
kick soccer ball & 60 & 11.44 $\pm$ \small1.68 & 12.36 $\pm$ \small1.93 & 10.94 $\pm$ \small1.95 & 12.90 $\pm$ \small2.03 & \textbf{8.96} $\pm$ \small0.88\\
kick soccer ball & 100 & 8.65 $\pm$ \small0.87 & 10.41 $\pm$ \small1.09 & 8.26 $\pm$ \small1.26 & 9.24 $\pm$ \small1.29 & \textbf{7.63} $\pm$ \small0.72\\
kick soccer ball & 200 & 6.12 $\pm$ \small0.43 & 7.63 $\pm$ \small0.64 & 6.23 $\pm$ \small0.77 & 6.97 $\pm$ \small0.91 & \textbf{5.83} $\pm$ \small0.38\\
\hline
Run & 30 & 15.71 $\pm$ \small2.89 & 15.92 $\pm$ \small2.16 & 14.17 $\pm$ \small2.00 & 15.75 $\pm$ \small2.51 & \textbf{10.00} $\pm$ \small0.97\\
Run & 60 & 13.49 $\pm$ \small2.71 & 13.89 $\pm$ \small1.82 & 10.63 $\pm$ \small1.42 & 12.70 $\pm$ \small1.66 & \textbf{8.95} $\pm$ \small0.79\\
Run & 100 & 9.85 $\pm$ \small1.53 & 10.36 $\pm$ \small0.69 & 8.96 $\pm$ \small0.74 & 9.33 $\pm$ \small0.88 & \textbf{7.34} $\pm$ \small0.48\\
Run & 200 & 6.44 $\pm$ \small0.78 & 7.24 $\pm$ \small0.33 & 6.42 $\pm$ \small0.25 & 6.75 $\pm$ \small0.54 & \textbf{5.89} $\pm$ \small0.34\\
\hline
Walk & 30 & 16.81 $\pm$ \small2.63 & 15.92 $\pm$ \small2.74 & 15.34 $\pm$ \small2.80 & 17.00 $\pm$ \small2.63 & \textbf{11.81} $\pm$ \small1.02\\
Walk & 60 & 14.47 $\pm$ \small2.02 & 14.31 $\pm$ \small2.10 & 13.75 $\pm$ \small1.92 & 15.41 $\pm$ \small2.16 & \textbf{10.32} $\pm$ \small0.88\\
Walk & 100 & 10.44 $\pm$ \small1.34 & 9.88 $\pm$ \small1.11 & 10.32 $\pm$ \small0.94 & 11.02 $\pm$ \small1.43 & \textbf{8.98} $\pm$ \small0.76\\
Walk & 200 & 7.43 $\pm$ \small0.94 & 6.11 $\pm$ \small0.79 & 6.03 $\pm$ \small0.47 & 6.89 $\pm$ \small0.69 & \textbf{5.34} $\pm$ \small0.43\\
\hline
\end{tabular}
\end{table*}
\subsection*{Sample $\gamma_{x}$:}
\begin{equation}
P(\gamma_{x}\mid -) \propto \prod_{j=1}^{M_{x}}\prod_{k=1}^{K}\mathcal{N}(d_{jk}^{x};0,\gamma_{x}^{-1})Gamma(\gamma_{x},a_{x},b_{x})
\end{equation}
It can be demonstrated that $\gamma_{x}$ can be drawn from a Gamma distribution
\begin{equation}
P(\gamma_{x}\mid -) \sim Gamma(a'_{x},b'_{x})
\end{equation}
where,
\begin{equation}
a'_{x} = a_{x} + \frac{M_{x}K}{2},\;\;\; b'_{x} = b_{x} + \frac{1}{2}\sum_{k=1}^{K}\sum_{j=1}^{M_{x}}(d_{jk}^{x})^{2}
\end{equation}

\subsection*{Sample $\gamma_{y}$:}
\begin{equation}
P(\gamma_{y}\mid -) \propto \prod_{j'=1}^{M_{y}}\prod_{k=1}^{K}\mathcal{N}(d_{j'k}^{y};0,\gamma_{y}^{-1})Gamma(\gamma_{y},a_{y},b_{y})
\end{equation}
It can be shown that $\gamma_{y}$ can be drawn from a Gamma distribution
\begin{equation}
P(\gamma_{y}\mid -) \sim Gamma(a'_{y},b'_{y})
\end{equation}
where,
\begin{equation}
a'_{y} = a_{y} + \frac{M_{y}K}{2},\;\;\; b'_{y} = b_{y} + \frac{1}{2}\sum_{k=1}^{K}\sum_{j'=1}^{M_{y}}(d_{j'k}^{y})^{2}
\end{equation}
\section*{Appendix B: Additional Results}
The average estimation accuracies (over 10 runs) together with the standard deviation for activities "Throw and catch football", "kick soccer ball", "Micheal Jacson style", "Run", and "Walk" are shown in Table \ref{table_example2}.
\section{Computational Complexity of the Proposed Method}
In this section, we consider the number of operations (addition and multiplication) for one sweep of our MCMC sampling method. The number of operations for sampling $\{z_{i}\}_{i=1}^{N}$ ($N$ is the number of training data) based on Eqs. 3 and 4 is in $O\big( NK(M_{x}+M_{y})\big)$ time ($K$ is the number of the dictionary atoms). The number of operations for sampling $\{s_{i}\}_{i=1}^{N}$ based on Eqs. 10 and 11 is in $O\big( NK(M_{x}+M_{y})\big)$ time. The time complexity for sampling $\{w_{k}\}_{k=1}^{K}$ is dominated by the computation of the inverse of $\Sigma_{w}$ which is in $O(N^{3})$ time. The number of operations for sampling $\{d_{k}^{x}\}_{k=1}^{K}$ and $\{d_{k}^{y}\}_{k=1}^{K}$ based on Eqs. 23 and 27 is in $O(NK^{2})$ time. The time complexity for sampling $\gamma_{x}, \gamma_y, \gamma_s$ and $\gamma_{xy}$ are in $O(KM_{x}), O(KM_{y}), O(KN)$ and $O(NK(M_x + M_y))$ time respectively. Hence, the computational complexity for one sweep of the MCMC method is approximately in $O(NK(M_x + M_y + K) + N^3)$ time.

\ifCLASSOPTIONcaptionsoff
  \newpage
\fi

\bibliographystyle{IEEEtran}
\bibliography{IEEEexample}

\end{document}